\pdfoutput=1

\documentclass[11pt]{article}

\usepackage[]{acl}

\usepackage{times}
\usepackage{latexsym}
\usepackage{graphicx}
\usepackage{amsmath}
\usepackage{url}
\usepackage{graphicx}
\usepackage{subfigure} 
\usepackage{multirow}
\usepackage{makecell}
\usepackage{colortbl}
\usepackage{amsfonts}

\usepackage{amssymb}
\usepackage{booktabs}
\usepackage{colortbl}
\usepackage{makecell}
\usepackage{verbatim}
\usepackage{latexsym}
\usepackage{graphicx,stackengine,scalerel}

\usepackage{stmaryrd}
\usepackage{helvet}
\usepackage{courier}
\usepackage{amsmath}
\usepackage{url}
\usepackage{multirow}
\usepackage{booktabs}
\usepackage{enumitem}
\usepackage{amssymb}
\usepackage{marvosym}
\usepackage{color}
\usepackage{caption}
\usepackage{titlesec}
\usepackage{floatrow}
\usepackage{caption}
\usepackage{rotating}
\usepackage{pifont}
\usepackage{xcolor}

\newfloatcommand{capbtabbox}{table}[][\FBwidth]

\definecolor{shallow_grey}{RGB}{167,194,203}
\definecolor{dark_grey}{RGB}{88,98,103}
\definecolor{light_green}{RGB}{66,195,183}
\definecolor{dark_blue}{RGB}{87,133,149}

\definecolor{dark_red}{RGB}{192,1,0}
\definecolor{dark_orange}{RGB}{255,147,2}
\definecolor{dark_yellow}{RGB}{255,213,121}

\definecolor{table_yellow}{RGB}{252,241,212}
\definecolor{table_green}{RGB}{198,233,230}
\definecolor{table_grey}{RGB}{197,200,202}

\usepackage[T1]{fontenc}

\usepackage[utf8]{inputenc}
\usepackage{verbatim}
\usepackage{enumitem}
\usepackage{array}

\usepackage{booktabs}
\usepackage{pgffor}

\usepackage{longtable} 

\usepackage{microtype}

\definecolor{darkgreen}{rgb}{0,0.35,0}

\definecolor{shallow_grey}{RGB}{167,194,203}
\definecolor{dark_grey}{RGB}{145,145,145}
\definecolor{light_green}{RGB}{66,195,183}
\definecolor{dark_blue}{RGB}{87,133,149}

\definecolor{dark_red}{RGB}{192,1,0}
\definecolor{dark_orange}{RGB}{255,147,2}
\definecolor{dark_yellow}{RGB}{255,213,121}

\definecolor{table_yellow}{RGB}{252,241,212}
\definecolor{table_green}{RGB}{198,233,230}
\definecolor{table_grey}{RGB}{197,200,202}

\title{How Reliable is Multilingual LLM-as-a-Judge?}

\author{Xiyan Fu \\
  Independent Researcher \\
  \texttt{fu@cl.uni-heidelberg.de} \\\And
  Anette Frank \\
  Dept. of Computational Linguistics \\
  Heidelberg University \\
  \texttt{frank@cl.uni-heidelberg.de} \\}

\author{Xiyan Fu$^{1}$  and Wei Liu$^{2}$ \\
  $^{1}$ Independent Researcher $^{2}$ Heidelberg Institute for Theoretical Studies gGmbH \\
  \texttt{xiyan.thea.fu@gmail.com | wei.liu@h-its.org}
}

\begin{document}
\maketitle
\begin{abstract}

LLM-as-a-Judge has emerged as a popular evaluation strategy, where advanced large language models assess generation results in alignment with human instructions. While these models serve as a promising alternative to human annotators, their reliability in multilingual evaluation remains uncertain. To bridge this gap, we conduct a comprehensive analysis of multilingual LLM-as-a-Judge. Specifically, we evaluate five models from different model families across five diverse tasks involving 25 languages. Our findings reveal that LLMs struggle to achieve consistent judgment results across languages, with an average Fleiss’ Kappa of approximately 0.3, and some models performing even worse. To investigate the cause of inconsistency, we analyze various influencing factors. We observe that consistency varies significantly across languages, with particularly poor performance in low-resource languages. Additionally, we find that neither training on multilingual data nor increasing model scale directly improves judgment consistency. These findings suggest that LLMs are not yet reliable for evaluating multilingual predictions. We finally propose an ensemble strategy which improves the consistency of the multilingual judge in real-world applications. 

\end{abstract}

\section{Introduction}
The success of various approaches based on neural networks has inspired the development of robust evaluation methods to track advances in the field of NLP \citep{sai2022survey, chang2024survey}. Evaluation aims to assess the quality and performance of NLP models, typically performed using evaluation metrics. Prior metrics vary depending on tasks and evaluation aspects, such as accuracy and F1-score for classification tasks, and BLUE \citep{papineni-etal-2002-bleu} and ROUGE \citep{lin2004rouge} for generation tasks. While these metrics benefit evaluations for various downstream tasks, their reliance on human-annotated references and n-gram matching limits their flexibility and effectiveness. With the development of deep learning, pre-trained language model-based evaluations are introduced, such as BLEURT \citep{sellam-etal-2020-bleurt} and BARTScore \citep{NEURIPS2021_e4d2b6e6}. They assess output quality by using pre-trained language model representations and generation probability. 

\begin{figure}
    \centering
    \includegraphics[width=\linewidth]{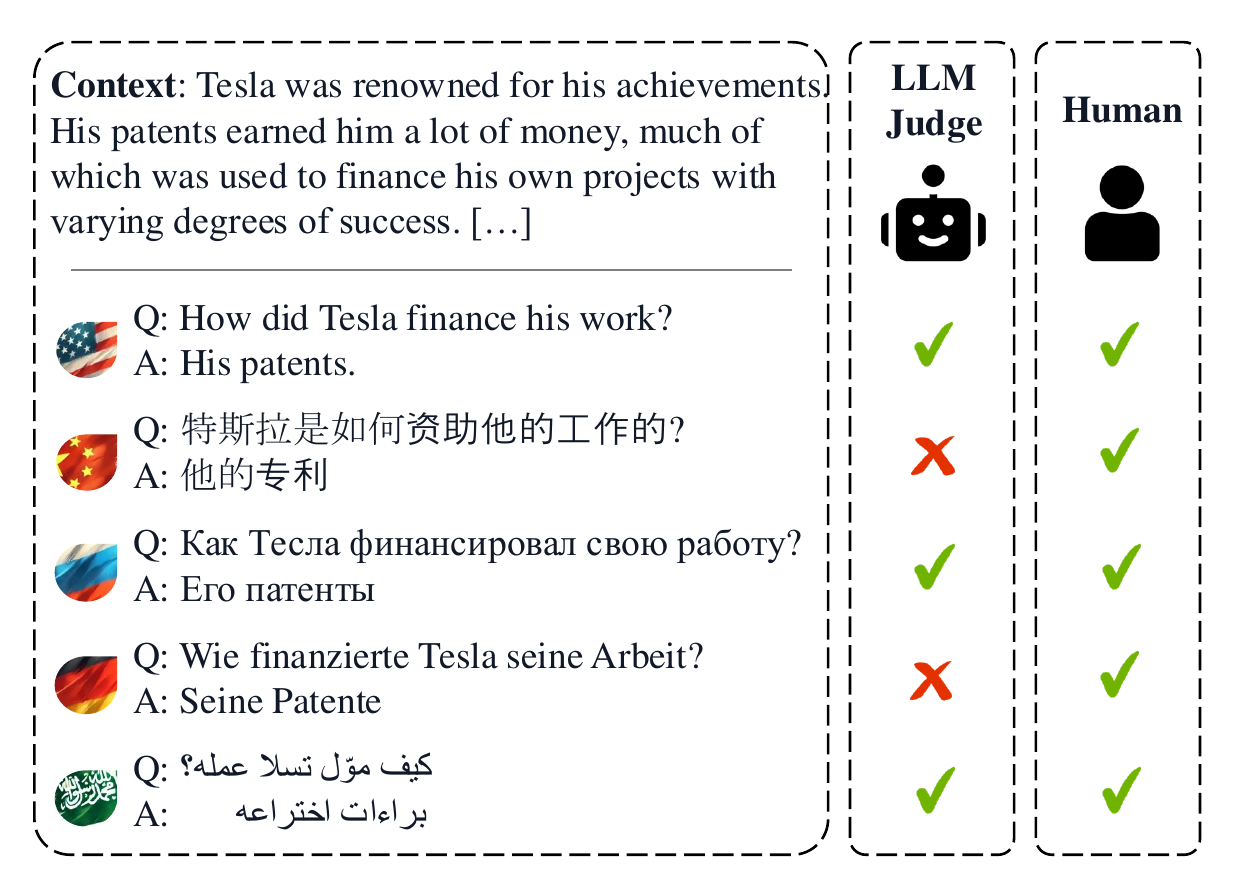}
    \caption{Inconsistency in multilingual LLM-as-a-Judge. Left part shows a multilingual Question Answering example. All question-answer pairs are parallel and perfectly aligned across languages. Human evaluators assess the results with uniform criteria. In contrast, LLM-as-a-Judge demonstrates inconsistency in its judgments, failing to maintain consistency across languages.}
    \label{fig:intro}
\end{figure}

To offer more efficient and powerful evaluation, some researchers propose LLM-as-a-judge \citep{zheng2023, li2024generation, gu2024survey}, which use powerful LLMs such as GPT4 \citep{achiam2023gpt} to evaluate generated response. \citet{fu-etal-2024-gptscore} defined evaluation schemes in the prompt template, and rely on existing LLMs as a judge to offer an evaluation. To avoid the high cost and potential data leakage, \citet{zhu2023judgelm} fine-tunes LLMs as their local evaluators. Existing works \citep{chiang-lee-2023-large} show that the result of LLM evaluation is consistent with the results obtained by expert human evaluation. These methods are subsequently applied to the evaluation of various tasks \citep{shen-etal-2023-large,fernandes-etal-2023-devil}. 

Given its superior performance, LLM-as-a-Judge has been extended to multilingual scenarios, where LLMs are expected to evaluate responses across different languages ~\citep{rau-etal-2024-bergen}. However, \textit{whether LLM-as-a-Judge is truly trustworthy for multilingual evaluation remains uncertain}. A reliable multilingual judge should be consistent, i.e., its judgments should depend on the content of the response rather than the language in which it is presented. Figure \ref{fig:intro} illustrates a multilingual Question Answering example, where question-answer pairs are parallel across various languages. A human annotator evaluates these responses consistently, without being influenced by language differences. 
To assess the reliability of multilingual LLM-as-a-Judge, we collect five datasets covering different tasks, each with parallel data across multiple languages. We evaluate five models and find that, despite achieving reasonable accuracy within each task, they all struggle to maintain consistent judgments across languages.

To further understand the factors affecting consistency, we analyze results across different dimensions. Notably, we observe that consistency scores for low-resource languages are significantly lower, even for multilingual LLMs designed for strong cross-lingual performance, such as Aya-Expanse \citep{dang2024ayaexpansecombiningresearch}. Furthermore, we find that the LLM's judgment consistency is influenced by its task-specific ability, highlighting the need to consider the alignment between the evaluation task and the model’s domain expertise. Overall, our findings shed light on the challenges of using LLM-as-a-Judge in multilingual settings and provide insights for future research on improving its reliability.

Our main contributions are as following:
\begin{itemize}
    \item We investigate the reliability of multilingual LLM-as-a-Judge by assessing its consistency across parallel multilingual data. Our findings reveal that LLMs struggle to provide consistent judgments across languages.
    \item We conduct a detailed analysis of factors that affect the LLM's consistency across languages. Experimental results show that multilingual LLM-as-a-Judge performs poorly in low-resource languages, and that the model's size and whether it undergoes multilingual training does not affect its consistency.
    \item We introduce an ensemble strategy to improve the consistency of the multilingual judge in real-world applications. 
\end{itemize}

\begin{figure}
    \centering
    \includegraphics[width=\linewidth]{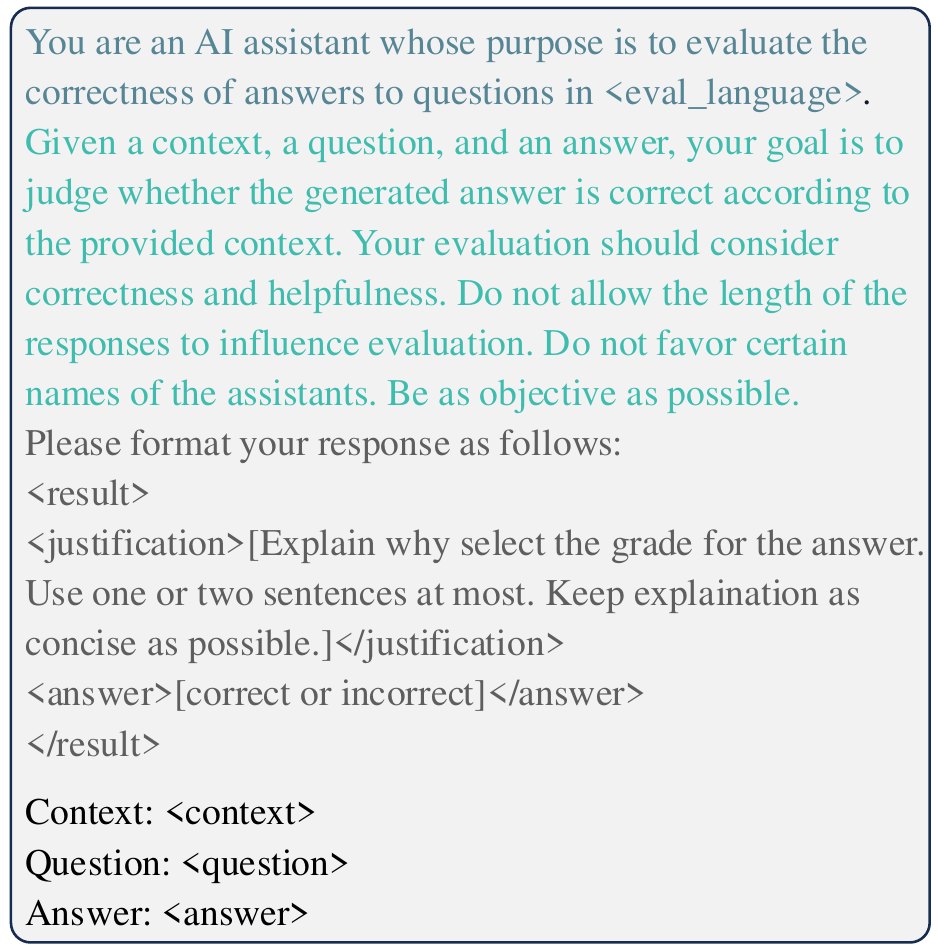}
    \caption{Prompt template for using LLM-as-a-Judge in a Question Answering task. Placeholders \textit{<eval\_language>, <context>, <question>, <answer>} are replaced by the input language, and its corresponding context, question and answer. The text in the prompt is color-coded to represent different sections: \raisebox{0.2\baselineskip}{\colorbox{dark_blue}{}} for role definition, \raisebox{0.2\baselineskip}{\colorbox{light_green}{}} for evaluation rubric, \raisebox{0.2\baselineskip}{\colorbox{dark_grey}{}} for output.}
    \label{fig:prompt}
\end{figure}

\section{Preliminary}
\label{sec2:preliminary}
\subsection{LLM-as-a-Judge}
\begin{table*}
    \centering
    \footnotesize 
    \resizebox{\columnwidth}{!}{
    \begin{tabular}{@{}llllc@{}} \toprule
        Dataset & Task &Answer Type  &Languages &Num \\ \midrule
        \makecell[l]{\textbf{XQuAD} \\ \citet{artetxe-etal-2020-cross}} &Question Answering &Extractive Span & \makecell[l]{English, German, Russian, Spanish, Chinese, Vietnamese,\\ Turkish, Greek, Romanian, Thai, Hindi} &1191\\ \midrule
        \makecell[l]{\textbf{MGSM} \\ \citet{shi2023language}} &Math Question Answering &Sentence & \makecell[l]{Spanish, French, German, Russian, Chinese, Japanese, \\Thai, Swahili, Bengali, Telugu} &250 \\ \midrule
        \makecell[l]{\textbf{WMT23} \\ \citet{kocmi-etal-2023-findings}} &Machine Translation &Sentence & \makecell[l]{English, Chinese, German, Japanese, Russian, Czech, \\ Ukrainian, Hebrew} & 196\\ \midrule
        \makecell[l]{\textbf{WikiLingua} \\ \citet{ladhak-etal-2020-wikilingua}} &Summarization &Sentence & \makecell[l]{English, Spanish, Castilian, Portuguese, French, German, \\ Russian, Italian, Indonesian, Dutch, Flemish, Arabic,  Chinese, \\Vietnamese, Thai, Japanese, Korean, Hindi, Czech, Turkish} & 142\\ \midrule
        \makecell[l]{\textbf{XDailyDialog} \\ \citet{liu-etal-2023-xdailydialog}} &Dialogue Generation &Sentence & \makecell[l]{English, Italian, Chinese, German} &996 \\ 
\bottomrule
    \end{tabular}}
    \caption{Datasets for multilingual LLM-as-a-Judge evaluation, all involving parallel data across provided languages. \textit{Num} indicates the number of data samples in one language.}
    \label{tab:dataset}
\end{table*}

LLM-as-a-Judge \citep{zheng2023} is a popular method that evaluates generated outputs without focusing on word-level matching or relying on highly cost human annotators. Instead, it uses powerful LLMs such as GPT4 \citep{achiam2023gpt} for evaluations covering multiple dimensions. Following \citet{gu2024survey}, we define a typical LLM-as-a-Judge as:
\begin{equation}
    \centering
    \begin{aligned}
        p \leftarrow {\rm{\tt LLM}}(C \otimes x)  \\
    \end{aligned}
    \vspace{-2pt}
\end{equation}
\noindent where $x$ is the input data awaiting evaluation, $C$ is the context of the input $x$, $\otimes$ is a combination operator that merges the input $x$ with the context $C$, ${\rm \tt{LLM}}$ is the model used for the judgment, and $p$ is the evaluation results from the whole LLM-as-a-Judge process. The context $C$ is usually a prompt template, containing (i) \textit{role definition}, which defines the task of the ${\rm \tt{LLM}}$; (ii) \textit{evaluation rubric}, which provides criteria and guidelines for evaluation; and (iii) \textit{output}, which regulates output formats and contents. Figure \ref{fig:prompt} shows a prompt example in the English Question Answering task.

Given the format of input $x$, LLM-as-a-Judge can be divided into two groups: (i) \textit{pointwise comparison} \citep{gao2023human}, where $x$ is a single candidate; (ii) \textit{pairwise comparison} \citep{fu-etal-2024-gptscore}, where $x$ is a pair involving candidate and reference. In this paper, we adopt pointwise evaluation for our experiments, as obtaining parallel multi-lingual candidates is challenging. Based on the format of the output, two judgment criteria exist: (i) \textbf{Yes / No} requires a binary judgment from LLMs, i.e., correct or incorrect. In this case, LLM-as-a-Judge solely focuses on accuracy. (ii) \textbf{Score} requires a discrete score from LLMs. Following \citet{chiang-lee-2023-large}, we define the score range as 1-5 given its superior evaluation performance. We use both criteria for the following experiments.

\subsection{Multilingual LLM-as-a-Judge}
In practice, multilingual evaluation is essential for assessing outputs across different languages, e.g., multilingual summarization. However, finding human annotators proficient in multiple languages is both challenging and costly. To address this, LLM-as-a-Judge is extended to Multilingual LLM-as-a-Judge. Compared to standard LLM-as-a-Judge, the input $x$ in this framework can appear in multiple languages beyond English. Figure \ref{fig:intro} illustrates an example. A reliable Multilingual LLM-as-a-Judge is expected to provide consistent judgments across parallel instances in different languages. 

%

\section{Experiment Setup}
\subsection{Models} 
We select five LLMs for experiments, including (i) GPT-3.5-turbo, GPT-4o-2024-08-06~\citep{OpenAI2024GPT4o}, since they are leading closed-source models which achieve State-ot-the-art results in a large range of NLP tasks; (ii) Llama-3.3-70b~\citep{dubey2024llama}, Qwen-2.5-72b~\citep{yang2024qwen2}, well known open source models; and (iii) Aya-expanse-32b~\citep{dang2024ayaexpansecombiningresearch}, multilingual specific model. The model is carefully trained using multilingual data arbitrage, multilingual preference optimization, and model merging methods, aiming to achieve robust multilingual capabilities. All the above models are commonly used as judges \citep{gu2024survey}.

\subsection{Tasks and Datasets}
Given our focus on exploring the consistency of LLM-as-a-judge in multilingual scenarios, we select datasets that contain \textit{parallel} data across all tested languages. The parallel structure of the dataset ensures that the input information remains identical across instances, with language being the only variable. The selected datasets cover a variety of NLP tasks, including Question Answering \citep{artetxe-etal-2020-cross}, Math Question Answering \citep{shi2023language}, Summarization \citep{ladhak-etal-2020-wikilingua}, Dialogue Generation \citep{liu-etal-2023-xdailydialog}, and Machine Translation \citep{kocmi-etal-2023-findings}, aiming to provide a comprehensive evaluation. Table \ref{tab:dataset} provides the details about these datasets.

\begin{table*}
    \centering
    \resizebox{\columnwidth}{!}{
    \begin{tabular}{llcccccccccc} \toprule
         &\multirow{2}{*}{Model}  & \multicolumn{2}{c}{XQuAD}  & \multicolumn{2}{c}{MGSM}  & \multicolumn{2}{c}{WMT23}  &  \multicolumn{2}{c}{XDailyDialog} & \multicolumn{2}{c}{WikiLingua} \\ \cmidrule(r){3-4} \cmidrule(r){5-6} \cmidrule(r){7-8} \cmidrule(r){9-10} \cmidrule(r){11-12}
         && Acc &FK & Acc &FK & Acc &FK & Acc &FK & Acc &FK  \\ \midrule
         \multirow{6}{*}{\rotatebox{90}{Yes / No}}&Aya-Expanse &96.86 & 0.2999  & 56.29 & 0.1895 & 92.64 &  0.1307 & 86.90 & 0.3812 & 89.87 & 0.3421 \\
         &Llama-3.3 &79.03 & 0.0748 & 64.25 & 0.0991 &53.57& 0.1463 &74.50&0.2425 &59.78 &0.2325 \\
         &Qwen-2.5  &93.47 & 0.3620&75.93 &0.2631 &92.42 &0.0775 &78.31 &0.3093 &67.68 &0.3531 \\
         &GPT-3.5 &97.67 &0.1399 &74.51 &0.1855 &94.17 &0.1327 &83.46 &0.2127 &56.14 &0.1748 \\
         &GPT-4o &92.04 &0.3694 &84.98 &0.2352 &85.88 &0.1691 &79.92 &0.3692 &65.57 &0.5424\\ 
         \midrule\midrule
         & &Avg &FK &Avg &FK &Avg &FK&Avg &FK &Avg &FK \\ \cmidrule(r){2-12}
         \multirow{5}{*}{\rotatebox{90}{Grade}}&Aya-Expanse &4.86 &0.2399 & 3.70 & 0.0260 & 4.58 & 0.1434 & 4.44 & 0.3049 & 4.46 & 0.1865 \\
         &Llama-3.3 & 4.64 &0.1558 & 3.64 & 0.1084 & 3.18 & 0.2082 & 3.73 & 0.1635 & 3.50 & 0.1412 \\
         &Qwen-2.5 &4.72&0.2926 &4.62 &0.0654 &4.79 &0.1471 &4.23 & 0.2602 &3.63 &0.2946  \\
         &GPT-3.5 &4.71 &0.0971 &3.57 &0.0660 &4.36 &0.1039 &4.06 &0.1240 &3.23 &0.0487 \\
         &GPT-4o &4.57 &0.3209 &3.66 &0.2041 &4.57 &0.1281 &4.24 &0.2405 &3.07 & 0.2803\\
         \bottomrule
    \end{tabular}}
    \caption{Performance of multilingual LLM-as-a-Judge across five datasets, evaluated on two settings: (i) \textit{Yes/No}, with binary evaluation accuracy (Acc), and (ii) \textit{Grade}, with average grade value (Avg) ranging from 1 to 5. Fleiss's Kappa (FK) is calculated for both settings to measure judgment consistency across parallel data.}
    \label{tab:main_rst}
\end{table*}

\subsection{Prompts}
For each test sample, we select ground truth as evaluated answers. This is to ensure precise parallel data alignment across all languages. Judgment instructions are then constructed as described in Section \ref{sec2:preliminary} and subsequently adapted into final prompts tailored for different models. Full templates are provided in the Appendix \ref{app:prompts}.

Existing studies \citep{sclar2024quantifying} have highlighted the critical role of prompt selection, as it significantly impacts final performance. Multilingual scenarios further amplify the challenges for LLM-as-a-Judge. Following \citep{ahuja-etal-2023-mega}, we adopt an English prompt with a specified target language indicated by `<eval\_language>' within the prompt, given its superior performance.

\subsection{Evaluation Metrics}
In this study, we focus on whether the performance of multilingual LLM-as-a-Judge varies significantly across parallel data in different languages. That is, whether it exhibits bias toward specific languages. Therefore, we select \textbf{Fleiss’ Kappa (FK)}, a statistical measure of inter-rater agreement for more than two raters, to measure the consistency of the  LLM-as-a-Judge results across languages. Here, we treat each model's output in a particular language as a rater's judgment.

While this study focuses on the consistency of LLM-as-a-Judge across languages, a truly excellent multilingual judge must also ensure accuracy. High consistency alone does not guarantee correctness, as it can result from uniformly incorrect judgments. To address this, we incorporate quality metrics to complement our evaluation:
(i) \textbf{Accuracy (Acc)}: For \textit{Yes/No} judgments we use accuracy to evaluate binary prediction. (ii) \textbf{Average Grade (AG)}: For \textit{Grade} judgment, we use average value to evaluate discrete grade prediction.
Notably, since we treat the ground truth as the predicted output to ensure precise parallel data alignment, the average accuracy and grade are expected to be 100\% and a score of 5, respectively.

\section{How does multilingual LLM-as-a-Judge perform?}

\subsection{Main Result}
\label{main_rst}
Table \ref{tab:main_rst} summarizes the performance of all multilingual LLMs-as-a-Judge across two judgment criteria: \textit{Yes/No} and \textit{Grade}. Based on Fleiss's Kappa metric, which measures consistency, GPT-4o achieves the highest performance, with a score of 0.5424 on WikiLingua for the \textit{Yes/No} criterion and 0.3209 on XQuAD for the \textit{Grade} criterion. However, these values remain far from the ideal consistency value of 1, and the Kappa scores of other models are even lower. This highlights that \textbf{even powerful LLMs struggle to act as fair and consistent multilingual judges}.

In addition, we observe significant variance in judgment consistency across different model groups. GPT-4o demonstrates superior Fleiss’ Kappa compared to other models, aligning with its state-of-the-art status in a wide range of NLP tasks. In contrast, GPT-3.5, a model from the same series as GPT-4o, exhibits notably lower consistency, with its Kappa scores typically around half of GPT-4o’s for both judgment criteria. However, despite GPT-4o attaining the highest Kappa consistency values, its judgment accuracy is not always the best. This contradicts the expectation that a strong judge should excel in both evaluation metrics. We speculate that this discrepancy arises from GPT-4o applying stricter evaluation standards rather than reflecting weaker performance. Such strictness makes it more challenging to achieve both high accuracy (exact correctness) or high ratings (score of 5) and high consistency simultaneously. Notably, we find that powerful open-source models, such as Qwen-2.5, achieve comparable performance to OpenAI models in multilingual judgment tasks. However, another open-source model, Llama-3.3, exhibits more limited performance. Furthermore, we experiment with Aya-Expanse, a multilingual LLM specifically fine-tuned on multilingual data. Despite this specialization, Aya fails to demonstrate noticeable improvements. This suggests that fine-tuning with multilingual data may not directly enhance a model’s ability to perform accurate multilingual judgments.


\begin{table}
    \centering
    \resizebox{\columnwidth}{!}{
    \begin{tabular}{l|ccccc} \toprule
         &XQuAD &MGSM &WMT23 &XDailyD &WikiL \\ \midrule
         YES/No&0 &0.6 &-0.5 &-0.3 &0.7 \\
         Grade & -0.2 &-0.5 &0 &0.5 &0.4 \\
         \bottomrule
    \end{tabular}}
    \caption{Spearman Correlation across five datasets for two judgment criteria: (i) Yes/ No, the correlation between accuracy and kappa; and (ii) Grade, the correlation between average value and kappa.}
    \label{tab:pearson}
\end{table}

\subsection{Consistency Result} 
To gain a deeper understanding of the performance of multilingual LLM-as-a-Judge, we further analyze the trends of Kappa consistency under the following settings:\par

\vspace{4pt}
\noindent \textbf{Acc / Avg VS. Kappa}. We analyze the relationship between prediction performance which is measured by Accuracy for \textit{Yes/No} and Average Score for \textit{Grade} and consistency measured by Kappa values. Specifically, we compute the Spearman correlation between accuracy (or average score) and Fleiss’ Kappa. Table \ref{tab:pearson} presents the results. We observe that the Spearman correlation varies inconsistently, depending on the evaluation tasks and judgment criteria. For the WikiLingua (WikiL) dataset, results show a positive correlation under two judgment criteria, 0.7 and 0.4 respectively. In contrast, other datasets present contrasting correlations, either positive or negative, two of them even 0. This suggests that \textbf{higher prediction accuracy does not necessarily imply greater judgment consistency}.

\par\vspace{4pt}

\begin{figure}
    \centering
    \includegraphics[width=1\linewidth]{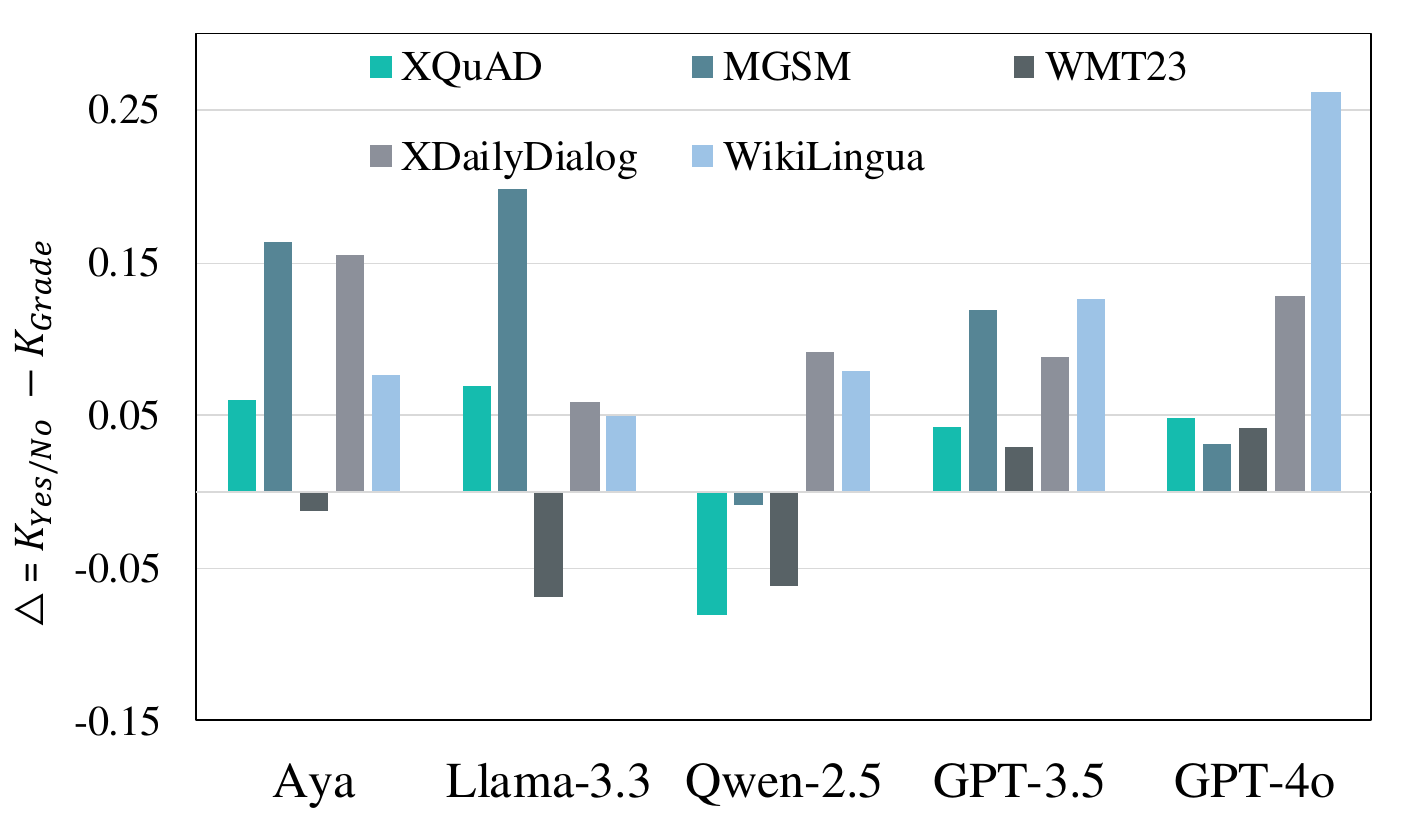}
    \caption{Fleiss Kappa value gap ($\Delta$) between \textit{Yes / No} and \textit{Grade} evaluation criteria of various multilingual LLM-as-a-Judge models.}
    \label{fig:gap}
\end{figure}

\noindent \textbf{Yes / No VS. Grade}. We further analyze the consistency, measured by Kappa values, across the two evaluation criteria: \textit{Yes / No} and \textit{Grade}. Specifically, we calculate the gap between the two criteria, defined as $\Delta$ = Kappa$_{Yes/No}$ - Kappa$_{Grade}$. Figure \ref{fig:gap} illustrates the gap across all datasets. We observe that most gap values are positive, i.e., consistency in \textit{Yes / No} evaluations is consistently higher than in \textit{Grade} evaluations. It indicates that grade judgment is more challenging than binary judgment. This result may be due to more options in the grade scale. In practice, limiting the options for LLM-as-a-Judge may enhance its effectiveness in applications that demand high multilingual consistency.

\begin{figure*}
    \centering
    \includegraphics[width=0.9\linewidth]{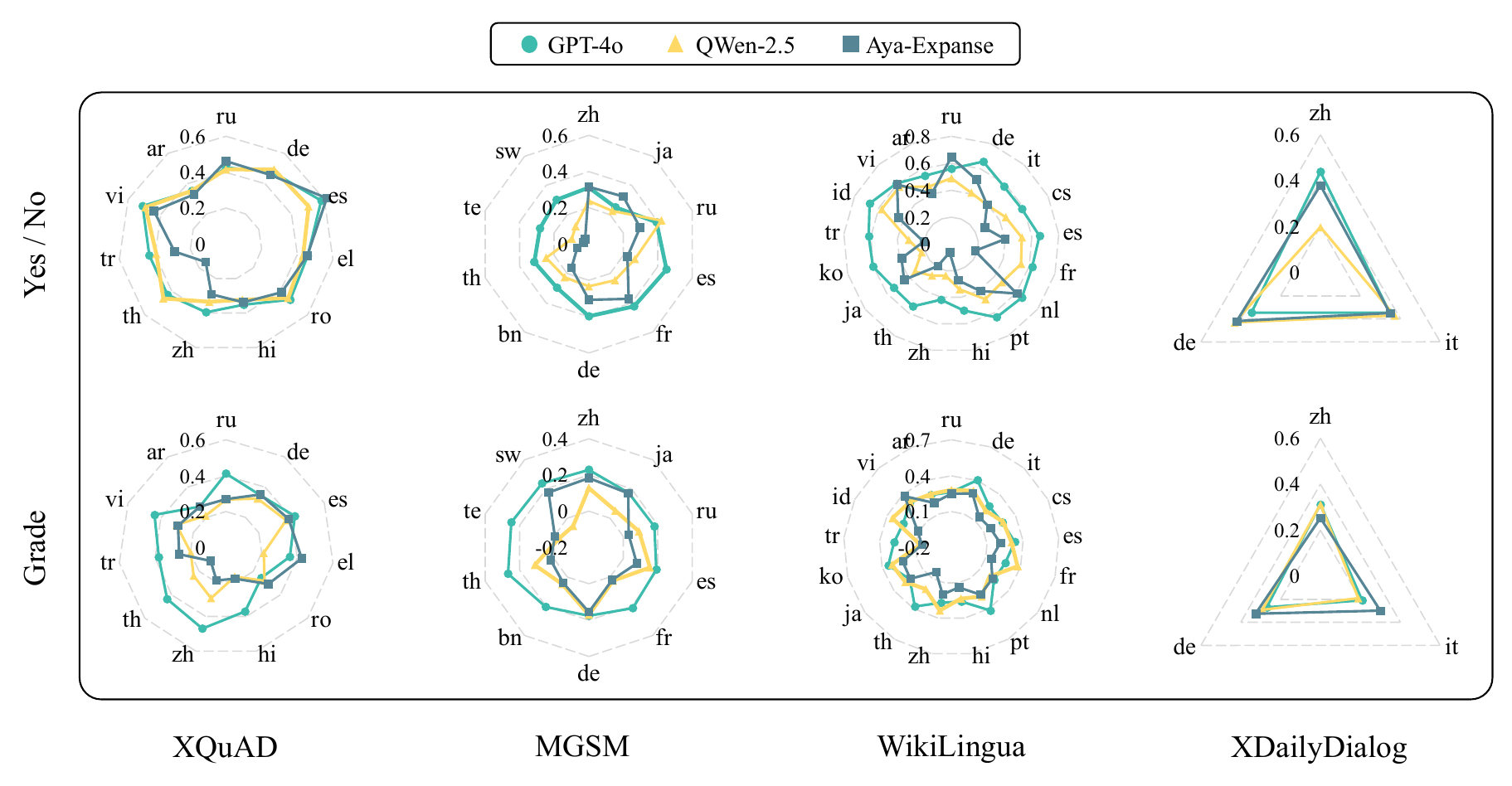}
    \caption{Consistency (Cohen’s Kappa) of LLMs’ judge results between English and other languages across four datasets and two judge criteria, \textit{Yes / No} and \textit{Grade}.}
    \label{fig:consistency_lang}
\end{figure*}
\label{lang_rst}

\section{What Factors cause inconsistency?}
To further understand the inferior consistency of multilingual LLM-as-a-Judge observed in the main results, we investigate potential causes in this section. 

\subsection{Correlation between Languages}
\label{lang_corre}
Existing works found that the training corpus of LLMs is usually dominated by English, so LLMs may perform strongly in English while being relatively weaker in other languages. Hence, we conduct an experiment to explore how close LLM-as-a-Judge performs in non-English languages compared to English. Specifically, we calculate the consistency (using Cohen's Kappa\footnote{Fleiss' Kappa is ignored as it works for more than 2 raters.}) between LLMs' judge results on English and those on other languages. We select three LLMs-GPT-4o, Qwen-2.5-70b, and Aya-Expanse-32b for experiments since they are a good mix of closed-source, open-source, and multilingual LLMs. Figure \ref{fig:consistency_lang} shows Cohen's Kappa results of four tasks\footnote{WMT23 is ignored here given experimented machine translation samples all contain English.} with two judge criteria.

The consistency radar charts for all tasks exhibit noticeable convex and concave patterns, indicating that consistency results with English vary across languages. Specifically, LLMs tend to show higher consistency with European languages. For example, on the XQuAD task, all judge results for Spanish and German show high consistency, with Cohen's Kappa values ranging from 0.30 to 0.61. This is likely due to (i) the LLMs’ training corpus containing more data in these languages, and (ii) their linguistic proximity to English (belonging to the same language family). In contrast, LLMs struggle with low-resource languages like Arabic (ar) and Telugu (te). For instance, on the MGSM task, the Cohen's Kappa value between Llama-3.3-70B judge results for Telugu and English is as low as 0.002. This trend persists even with Aya-Expanse-32B, a multilingual LLM with strong capabilities. These findings suggest that \textbf{we must be cautious when using LLM evaluation results for low-resource languages, as they may be unreliable}.



\subsection{Impact of the judged task}
\label{sec:kappa_tasks}
Figure \ref{fig:consistency_lang} also shows that the radar charts vary significantly across different tasks. Specifically, on the XQuAD task, the consistency between LLMs' judge results on English and other languages generally ranges from 0.2 to 0.4, with GPT-4o and Qwen-2.5-72b performing the best. In contrast, the consistency results on the MGSM task drop to around 0.2, and the results of Qwen-2.5-72b and Aya-Expanse-32b for some languages are even close to 0 in terms of consistency with the results in English. However, on the WikiLingua task, the consistency results (in the \textit{Yes}/\textit{No} setting) climb to as high as 0.8. This suggests that \textbf{when choosing a multilingual LLM-as-a-Judge for tasks, one should consider the LLM's task-related capabilities}. The results of Aya-Expanse-32b confirm this to some extent. Aya-Expanse-32b is an LLM carefully trained to aim for strong multilingual capacities. However, surprisingly, it shows the worst consistency between judge results on English and other languages, especially on the MGSM task. We speculate that this is because Aya-Expanse-32b has not been primarily trained to solve reasoning and mathematical problems. This leads to its poor performance when evaluating the MGSM task, which consists of mathematical questions. Furthermore, we find that GPT-4o exhibits the best consistency across all tasks and languages, indicating its superiority in building multilingual LLM-as-a-Judge.



\subsection{Prompt Design}
\label{sec:prompt}

Existing research \citep{sclar2024quantifying} has identified prompt design as a key factor in LLM-as-a-Judge performance. Therefore, we investigate how prompt design influences multilingual judgment consistency. As described in Section \ref{sec2:preliminary}, the instruction prompt in this work consists of three components: role definition, evaluation rubric, and output format. Since the role definition of LLM-as-a-Judge is generally static, our experiments primarily focus on the latter two components. For the \textit{evaluation rubric}, we tested: (i) a general rubric, which defines a grading scale with simplified descriptions for evaluation, and (ii) a specific rubric, which defines a grading scale where each grade is accompanied by detailed rules and explanations. For the \textit{output format}, we tested: (i) prediction only, where LLMs output a simple binary prediction or evaluation grade, and (ii) prediction with explanation, where LLMs provide both the prediction and the reasoning behind their judgment. Table \ref{tab:prompt} shows the results for different prompt designs by combining these two factors.

\begin{table}
    \centering
    \resizebox{\columnwidth}{!}{
    \begin{tabular}{cl|cccc} \toprule
         \multirow{2}{*}{ID} &\multirow{2}{*}{Prompt}  &\multicolumn{2}{c}{XQuAD} &\multicolumn{2}{c}{WMT23} \\ \cmidrule(r){3-4} \cmidrule(r){5-6} 
         &&Avg& Kappa &Avg& Kappa\\  \midrule
         \ding{172}&\makecell[l]{rubric: general \\ out: prediction} &4.66	&0.2517	&4.55 &0.1133 \\ \midrule
         \ding{173}&\makecell[l]{rubric: general \\ out: prediction + explaination} &4.57	& 0.3209 &4.57 &0.1281 \\ \midrule
         \ding{174}&\makecell[l]{rubric: specific \\ out: prediction} &4.63	&0.2145	&4.57 &0.1145\\ \midrule
         \ding{175}&\makecell[l]{rubric: specific \\ out: prediction + explaination} &4.67	&0.2239 &4.63  &0.1196\\ \bottomrule
    \end{tabular}}
    \caption{Variation of Accuracy (Acc) and Fleiss Kappa with different prompt templates for \textit{Grade} judgment of GPT-4o. \textit{rubric} and \textit{out} represent evaluation guideline and output requests as shown in Section \ref{sec2:preliminary}.}
    \label{tab:prompt}
\end{table}

By comparing the consistency values (Kappa) between prompts with and without explanation generation (i.e., \ding{172} vs. \ding{173} and \ding{174} vs. \ding{175}), we observe that prompts with explanation generation consistently achieve superior results. This indicates that \textbf{generating explanations to support judgments can enhance evaluation consistency across all languages}. Additionally, we compare prompts with general and specific rubrics (i.e., \ding{172} vs. \ding{174} and \ding{173} vs. \ding{175}). Interestingly, we find that providing specific rules does not always improve consistency. We speculate that this may be because LLMs are already familiar with commonly used tasks, making very specific rubrics unnecessary in certain cases.

\subsection{Model Scale}
\label{sec:scale}
\begin{figure}
    \centering
    \includegraphics[width=1\linewidth]{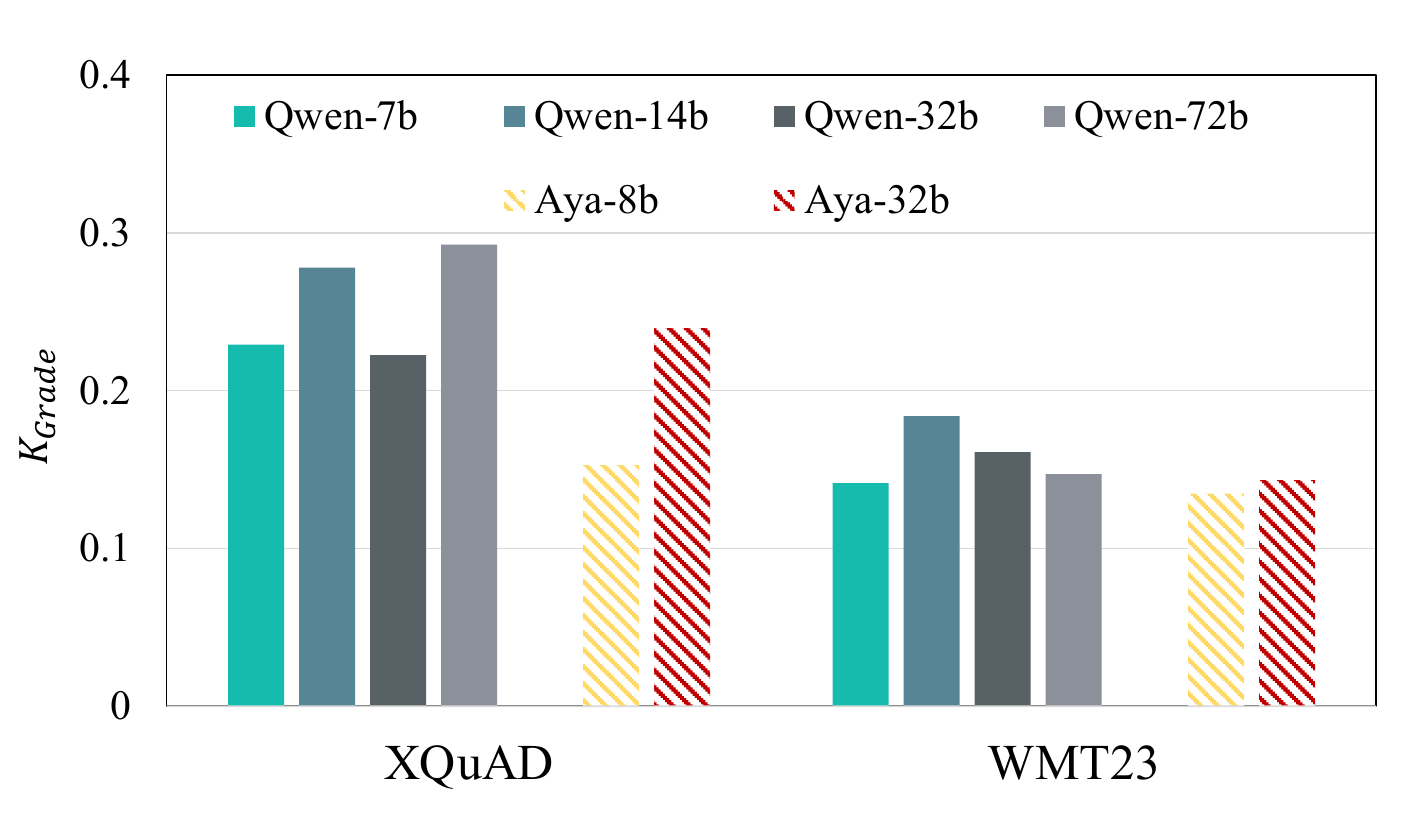}
    \caption{Variation of Fleiss Kappa for \textit{Grade} judgment (K$_{Grade}$) across Qwen-2.5 and Aya-Expanse in different model scale.} 
    \label{fig:scale}
\end{figure}

We further investigate whether the scale of LLMs affects inconsistency across languages. Specifically, we examine the open-access model Qwen-2.5, which ranges from 7 billion to 72 billion parameters, and the multilingual-specific model Aya-Expanse, which ranges from 7 billion to 32 billion parameters. Table \ref{fig:scale} presents the results.


For Qwen-2.5 across different model scales, we do not observe any consistent trend. On the WMT23 dataset, the 14-billion-parameter Qwen-2.5 model even achieves higher consistency compared to the 72-billion version. Additionally, while the 32-billion Aya-Expanse outperforms its smaller counterparts, its improvement on WMT23 remains limited. These findings suggest that increasing the model scale does not directly lead to enhanced consistency in multilingual LLM-as-a-Judge.

\section{How to choose a Judge in the wild?}
Existing results show that Multilingual LLM-as-a-Judge exhibits varying consistency across different languages and tasks. This raises a natural question: \textit{How can we choose a suitable LLM-as-a-Judge for real-world applications to ensure relatively consistent evaluations across languages}? Table \ref{tab:main_rst} indicates that GPT-4o generally achieves the highest consistency, making it an ideal choice. However, its high cost and potential risk of data leakage pose challenges. To address this, we propose an Ensemble strategy that leverages a majority vote among open-source LLMs for judgment, inspired by \citet{verga2024replacing,raina-etal-2024-llm}.

Specifically, we conduct experiments using three open-source LLMs: Llama-3.3-70B, Qwen-2.5-72B, and Aya-Expanse-32B, taking their majority vote as the final prediction. The ensemble results (Ens) are shown in Table \ref{tab:ensemble}. For comparison, we also report the minimum value (Min) among the three models, representing the worst-case scenario when the least reliable judge is unknowingly selected. Furthermore, we compute the gap between the ensemble results and the minimum value, denoted as $\Delta = Ens - Min$, which reflects the improvement over the worst-case performance. As shown in Table \ref{tab:ensemble}, most gap values are positive, except for -0.0046 in WMT23 and -0.0142 in another case. Given that other improvements are generally above 0.1, we conclude that \textbf{the ensemble strategy can enhance consistency in real-world applications where the least reliable LLM might be unknowingly chosen}.

\begin{table}
    \centering
    \resizebox{\columnwidth}{!}{
    \begin{tabular}{lccccc} \toprule
         &XQuAD &MGSM &WMT23 &XDailyD &WikiL \\ \midrule
         \multicolumn{2}{l}{\textit{Yes / No} :} \\
         \multicolumn{1}{l|}{Min} &0.0748 &0.0991 &0.0775 &0.2425 &0.2325 \\
         \multicolumn{1}{l|}{Ens} &0.3227 &0.2162 &0.0729 &0.4053 &0.4217 \\  \cmidrule(r){2-6}
         \multicolumn{1}{l|}{$\Delta$} &0.2479 &0.1171 &-0.0046 &0.1628 &0.1892 \\ \midrule
         \multicolumn{2}{l}{\textit{Grade} :} \\
         \multicolumn{1}{l|}{Min} &0.1558 &0.0654 &0.1434 &0.1635 &0.1412 \\
         \multicolumn{1}{l|}{Ens} &0.2617 &0.0512 &0.2078 &0.1675 &0.2931  \\  \cmidrule(r){2-6}
         \multicolumn{1}{l|}{$\Delta$} &0.1059 &-0.0142 &0.0644 &0.0040 &0.1519 \\
         \bottomrule
    \end{tabular}}
    \caption{Ensemble results (Ens) of Aya, QWen, and Llama. \textit{Min} indicates the minimum consistency of the above three models. $\Delta$ shows the gap between ensemble results and minimum value, i.e., $\Delta$ = Ens - Min.}
    \label{tab:ensemble}
\end{table}

\section{Related Work}
\subsection{LLM-as-a-judge} 
With the remarkable performance of LLMs, researchers have increasingly leveraged them to evaluate generation results in alignment with human instructions \citep{zheng2023}, known as \textit{LLM-as-a-judge}. To apply LLM-as-a-judge, it is common to start using In-Context Learning \citep{brown2020language} methods with advanced LLMs, such as GPT-4 \citep{achiam2023gpt}. \citet{li2024generation} categorized evaluation prompts into two primary groups: (i) \textit{pairwise comparison}, where an LLM is given two candidates along with context to determine which response is superior \citep{gao2023human}; and (ii) \textit{pointwise evaluation}, where an LLM assesses a single candidate based on specified evaluation criteria \citep{fu-etal-2024-gptscore}. To further enhance LLMs’ judging capabilities, other line works apply preference learning techniques \citep{wang2024self,wu2024meta} and fine-tuning mechanism \citep{zhu2023judgelm}. These methodologies have been extensively applied across various tasks, including summarization \citep{shen-etal-2023-large, wang-etal-2023-chatgpt}, translation \citep{kocmi-federmann-2023-large, fernandes-etal-2023-devil}, and written discourse coherence \citep{naismith-etal-2023-automated}. The widespread adoption of LLM-as-a-judge raises questions about its reliability and effectiveness. Addressing this, \citet{chiang-lee-2023-large} validated its efficacy by comparing evaluation outcomes from human judges and LLM-as-a-judge, further highlighting its potential to significantly enhance efficiency. As a complement to existing research, we focus on LLM-as-a-Judge in multilingual scenarios. Recently, \citet{hada-etal-2024-large} observed that LLM evaluators tend to assign higher scores in multilingual settings compared to human annotations. Unlike their work, we investigate the reliability of LLMs-as-a-Judge by examining their \textit{consistency} across different languages using parallel multilingual data.

\subsection{Bias} Despite the success of LLM-based evaluators, there have been studies showing that they have some biases \citep{zheng2023}. One well-explored bias is \textit{position bias} \citep{wang-etal-2024-large-language-models-fair,shi2024judging} that the evaluation ranking of candidate responses can be easily hacked by altering their order of appearance in the context. \citet{saito2023verbosity,park-etal-2024-offsetbias} introduced \textit{length bias} that LLMs prefer more verbose answers even if they have similar qualities, and \textit{authority bias} that LLMs favor responses with specific details, e.g., citation of authoritative sources. To address the effect of length, \citet{dubois2024lengthcontrolled} introduced a debiasing strategy given regression-based adjustments for observational causal inference. Beyond these superficial biases, \citet{park-etal-2024-offsetbias} identified four additional biases, such as familiar knowledge bias which refers to a preference for responses describing commonly encountered knowledge in real-world data. \citet{ye2024justice} highlighted the self-enhancement bias, where LLMs tend to favor responses generated by themselves. Instead, we evaluate biases in LLM-as-a-Judge with a focus on multilingual bias. We found that LLM-as-a-Judge struggles to provide consistent judgments across parallel inputs in different languages, with performance being particularly inferior for low-resource languages.

\section{Conclusion}
In this paper, we conduct an in-depth analysis of multilingual LLM-as-a-Judge, focusing on the consistency of its judgments across parallel data in different languages. Our results show that even advanced LLMs struggle with consistent judgment, exhibiting significant variance across languages. Moreover, neither larger model scales nor specific multilingual training improves judgment reliability. 
Our comprehensive analysis provides novel insights into multilingual LLM-as-a-Judge.

\section{Limitation}
For LLM-as-a-Judge, we focus on pointwise judgment, as obtaining parallel multilingual incorrect candidates is challenging. This limits its applicability in real-world scenarios. An interesting avenue for future work would be to construct a parallel pairwise corpus for evaluation.


Moreover, due to GPU constraints, we evaluate only open-access models up to approximately 70 billion parameters. Future work will explore judgments from larger LLMs.

\bibliography{anthology,custom}
\bibliographystyle{acl_natbib}
\clearpage

\appendix
\onecolumn 

\section{Experiment Details}
\subsection{Prompts}
\label{app:prompts}

\begin{longtable}{lp{15cm}}
    \toprule
    \textbf{Task} & \textbf{Prompt} \\ 
    \hline
    \endfirsthead
    
    \hline
    \textbf{Task} & \textbf{Prompt} \\
    \hline
    \endhead

    &\textit{output\_format\_Yes/No}: Please format your response as follows: <result><justification>[Explain why select the grade for the answer. Use one or two sentences at most. Keep explanation as concise as possible.]</justification><answer>[correct or incorrect]</answer></result> \\ \cmidrule(r){2-2}
         &\textit{output\_format\_Grade}: Please format your response as follows: <result><justification>[Explain why select the grade for the answer. Use one or two sentences at most. Keep the explanation as concise as possible.]</justification><answer>[a grade from {1, 2, 3, 4, 5}]</answer></result> \\ \midrule\midrule
         \multirow{10}{*}{XQuAD}& \textit{input}: Context: <context>; Question: <question>; Answer: <answer> \\ \cmidrule(r){2-2}
         & \textit{prompt\_Yes/No}: You are an AI assistant whose purpose is to evaluate the correctness of answers to questions in EVALUATION\_LANGUAGE. Given a context, a question, and an answer, your goal is to judge whether the generated answer is correct according to the provided context. Your evaluation should consider correctness and helpfulness. Do not allow the length of the answer to influence your evaluation. Be as objective as possible. <input> <output\_format\_Yes/No> \\ \cmidrule(r){2-2}
         & \textit{prompt\_Grade}: You are an AI assistant whose purpose is to evaluate the correctness of answers to questions in EVALUATION\_LANGUAGE. Given a context, a question, and an answer, your goal is to rate the generated answer on a scale from 1 (worst) to 5 (best). Your evaluation should consider correctness and helpfulness. Do not allow the length of the answer to influence your evaluation. Be as objective as possible. <input> <output\_format\_Grade> \\ \midrule
         \multirow{10}{*}{MGSM}& \textit{input}: Question: <question>; Answer: <answer> \\ \cmidrule(r){2-2}
         & \textit{prompt\_Yes/No}: You are an AI assistant whose purpose is to evaluate the correctness of answers to questions in EVALUATION\_LANGUAGE. Given a question and an answer, your goal is to judge whether the generated answer is correct. Your evaluation should consider correctness and helpfulness. Do not allow the length of the answer to influence your evaluation. Be as objective as possible. <input> <output\_format\_Yes/No> \\ \cmidrule(r){2-2}
         & \textit{prompt\_Grade}: You are an AI assistant whose purpose is to evaluate the correctness of answers to questions in EVALUATION\_LANGUAGE. Given a question and an answer, your goal is to rate the generated answer on a scale from 1 (worst) to 5 (best). Your evaluation should consider correctness and helpfulness. Do not allow the length of the answer to influence your evaluation. Be as objective as possible. <input> <output\_format\_Grade> \\ \midrule
         \multirow{10}{*}{WMT23}& \textit{input}: Source: <source>; Target: <target> \\ \cmidrule(r){2-2}
         & \textit{prompt\_Yes/No}: You are an AI assistant whose purpose is to evaluate the correctness of machine translation from English to EVALUATION\_LANGUAGE. For each pair of sentences, evaluate whether the translated sentence is correct. Your evaluation should consider correctness and helpfulness. Do not allow the length of the answer to influence your evaluation. Be as objective as possible. <input> <output\_format\_Yes/No> \\ \cmidrule(r){2-2}
         & \textit{prompt\_Grade}: \textit{prompt\_Yes/No}: You are an AI assistant whose purpose is to evaluate the correctness of machine translation from English to EVALUATION\_LANGUAGE. For each pair of sentences, evaluate the quality of the translated sentence on a scale from 1 (worst) to 5 (best). Your evaluation should consider correctness and helpfulness. Do not allow the length of the answer to influence your evaluation. Be as objective as possible. <input> <output\_format\_Yes/No>  \\ \midrule
         & \\
         & \\
         & \\
         \multirow{10}{*}{WikiL}& \textit{input}: Document: <document>; Summarization: <summarization> \\ 
         \cmidrule(r){2-2}
         & \textit{prompt\_Yes/No}: You are an AI assistant whose purpose is to evaluate the correctness of summarization in EVALUATION\_LANGUAGE. Given a document, and a summary, your goal is to judge whether the generated summary is correct according to the provided document. Your evaluation should consider correctness and helpfulness. Do not allow the length of the answer to influence your evaluation. Be as objective as possible. <input> <output\_format\_Yes/No> \\ \cmidrule(r){2-2}
         & \textit{prompt\_Grade}: \textit{prompt\_Yes/No}: You are an AI assistant whose purpose is to evaluate the correctness of summarization in EVALUATION\_LANGUAGE. Given a document, and a summary, your goal is to rate the generated summary on a scale from 1 (worst) to 5 (best). Your evaluation should consider correctness and helpfulness. Do not allow the length of the answer to influence your evaluation. Be as objective as possible. <input> <output\_format\_Yes/No>  \\ \midrule
         \multirow{10}{*}{XDailyD}& \textit{input}: Dialog: <dialog>; Next Utterance: <next\_utterance> \\ 
         \cmidrule(r){2-2}
         & \textit{prompt\_Yes/No}: You are an AI assistant whose purpose is to evaluate the correctness of dialogue generation in EVALUATION\_LANGUAGE. Given a dialog, your goal is to judge whether the generated next utterance is correct. Your evaluation should consider correctness and helpfulness. Do not allow the length of the answer to influence your evaluation. Be as objective as possible. <input> <output\_format\_Yes/No> \\ \cmidrule(r){2-2}
         & \textit{prompt\_Grade}: \textit{prompt\_Yes/No}: You are an AI assistant whose purpose is to evaluate the correctness of dialogue generation in EVALUATION\_LANGUAGE. Given a dialog, your goal is to rate the generated utterance on a scale from 1 (worst) to 5 (best). Your evaluation should consider correctness and helpfulness. Do not allow the length of the answer to influence your evaluation. Be as objective as possible. <input> <output\_format\_Yes/No>  \\
    \bottomrule
    \caption{Prompts of Multilingual LLM-as-a-Judge for various tasks.} \\
\end{longtable}

\end{document}